%% file: main.tex
\documentclass[letterpaper]{article} 
\usepackage[preprint]{aaai2027}  
\usepackage[hyphens]{url}  
\usepackage{graphicx} 
\usepackage{natbib}  
\usepackage{caption} 
\usepackage{algorithm}
\usepackage{algorithmic}
\usepackage{xspace}
\usepackage{amsmath}
\usepackage{bibentry}
\usepackage{placeins}
\usepackage{newfloat}
\usepackage{listings}
\DeclareCaptionStyle{ruled}{labelfont=normalfont,labelsep=colon,strut=off} 
\floatstyle{ruled}
\newfloat{listing}{tb}{lst}{}
\floatname{listing}{Listing}

\usepackage{booktabs}
\usepackage{tabularx}
\newcolumntype{C}[1]{>{\centering\arraybackslash}p{#1}}
\usepackage{colortbl}
\definecolor{besthl}{HTML}{DCEAF7} 
\definecolor{oursrow}{HTML}{F2F6FB} 

\newcommand{\tool}{\textsc{TACT}\xspace}
\newcommand{\tactcorpus}{TACTCorpus\xspace}
\newcommand{\tactbench}{TACTBench\xspace}
\newcommand{\tactutor}{TACTutor\xspace}
\newcommand{\eg}{\textit{e.g.,~}}
\title{\tool: Taxonomy-Aligned Post-Training for Pedagogically \\Adaptive English Tutoring}
\author{
    Dongjie Yang$^{1,*}$,
    Siyan Lin$^{2,*}$,
    Leixian Shen$^{1}$,
    Rui Sheng$^{1}$,
    Huamin Qu$^{1}$,
    Zixin Chen$^{1,\dagger}$
}
\affiliations{
    $^{1}$The Hong Kong University of Science and Technology, Hong Kong SAR, China\\
    $^{2}$Carnegie Mellon University, Pittsburgh, PA, USA\\
    {\small $^{*}$Equal contribution. $^{\dagger}$Corresponding author: zchendf@connect.ust.hk}
}

\begin{document}

\maketitle

\input{sections/00-abstract}


\input{sections/01-introduction}

\input{sections/02-theoretical-framing}

\input{sections/03-taxonomy-and-data-construction}

\input{sections/04-post-training-method}
\input{sections/05-evaluation-and-results}
\input{sections/06-discussion}
\input{sections/07-conclusion-and-limitations}

\input{sections/07-ethical-statement}
\bibliography{aaai2027}
\clearpage
\appendix
\setcounter{figure}{0}
\setcounter{table}{0}
\setcounter{listing}{0}
\setcounter{algorithm}{0}
\setcounter{equation}{0}
\renewcommand{\thefigure}{\arabic{figure}}
\renewcommand{\thetable}{\arabic{table}}
\renewcommand{\thelisting}{\arabic{listing}}
\renewcommand{\thealgorithm}{\arabic{algorithm}}
\renewcommand{\theequation}{\arabic{equation}}

\input{sections/08-appendix}



\end{document}

%% file: sections/00-abstract.tex
\begin{abstract}
\begingroup\catcode`\%=12
Large language models (LLMs) are increasingly used to provide conversational practice for English-as-a-second-language (ESL) learners. 
Effective ESL tutoring, however, requires more than fluent response generation: a tutor must select an appropriate pedagogical action (\eg eliciting self-correction, providing a hint, or recasting an utterance) based on learner behavior and dialogue context.
Human-tutoring research offers principles for adaptive support, but they are often task-specific and remain insufficiently integrated into LLM-based ESL tutor training and evaluation. 
We present \tool (\textbf{T}axonomy-\textbf{A}ligned \textbf{C}onversational \textbf{T}utor), a human-grounded framework for post-training and evaluating pedagogically adaptive ESL tutors. Drawing on established literature, we develop two complementary taxonomies: the Tutor-Strategy Taxonomy, comprising 13 tutor response strategies, and the Student-Move Taxonomy, characterizing learner behavior through two dimensions: student-move type and move status. Using these taxonomies, we construct \tactcorpus, which enriches 260 authentic teacher--student conversations with 32,379 annotations and quality-controlled augmented training data. We then post-train Qwen3.5-4B through supervised fine-tuning followed by taxonomy-aligned Group Relative Policy Optimization, producing \tactutor and optimizing it for scaffolding quality rather than reference imitation alone.
On \tactbench, a strategy-balanced diagnostic benchmark comprising 78 authentic tutoring contexts, \tactutor improves over its backbone by 20.30\% and outperforms all evaluated proprietary baselines under the same protocol, while maintaining backbone performance on established external educational benchmarks; in a blinded study with 50 learners, it also receives the highest overall mean rating among the evaluated tutors. We release the data, benchmark, and model weights. \tool provides an open foundation for translating human language-tutoring strategies into pedagogically adaptive agents, broadening access to personalized language-learning support at scale.
\endgroup
\end{abstract}

%% file: sections/01-introduction.tex
\section{Introduction}

Developing conversational proficiency in a second language benefits from frequent, meaningful interaction and timely feedback, but sustained one-to-one tutoring is costly and difficult to scale. Large language models (LLMs) offer a promising means of broadening access to conversational practice for learners of English as a second language (ESL).  Effective ESL tutoring, however, requires more than fluent dialogue: a tutor must decide whether, when, and how to intervene based on the learner's contribution and the surrounding interactional context~\cite{pena2024learning}. In language-learning interaction, each learner utterance both advances the conversation and reveals aspects of the learner's developing linguistic competence~\cite{seedhouse2004interactional}.  The tutor must therefore maintain engagement and learner agency while deciding whether a linguistic or pragmatic problem warrants attention and which instructional action is appropriate.

Existing LLM tutors often elicit pedagogical behavior through inference-time control. General tutoring systems use prompting or specialized configurations to encourage guided questioning, hinting, and Socratic reasoning, while language tutors may additionally condition responses on learner proficiency or strategically use the learner's first language~\citep{openai2023teaching,openai2023khan,openai2025study, anthropic2025education,almasi2025alignment,liu2024codeswitching}.
Within ESL, prior work has explored prompted response generation and separate tutor-act planning~\citep{vanzo2025gpt,kwon2024biped}. Although these studies demonstrate the value of explicit pedagogical control, two limitations remain: First, such control often remains an inference-time scaffold rather than a decision policy internalized by an open response model. This dependence complicates the development of compact, self-contained tutors for local deployment on personal devices, where inference cost, latency, and the privacy of learner conversations are important considerations. Second, pedagogically annotated data from authentic ESL tutoring interactions remain limited, and a unified operational framework that links a learner's dialogue move and repair status to the tutor strategy that follows is still lacking. 

Reliable learner-contingent pedagogical decision-making in ESL interaction therefore remains an unresolved challenge for compact open models. Without this capability, a model may remain seemingly helpful while making an inappropriate pedagogical decision—for example, correcting a largely acceptable utterance rather than acknowledging it to encourage the learner and sustain conversational flow, or supplying the answer while the learner is still reflecting on an incorrect but repairable attempt rather than offering a hint that supports continued reflection and preserves learner agency.

To address this, we introduce \tool\footnote{Model weights and data samples are available at \url{https://huggingface.co/Taxonomy-Aligned-Conversational-Tutor}.} (\textbf{T}axonomy-\textbf{A}ligned \textbf{C}onversational \textbf{T}utor), a framework for training and evaluating pedagogically adaptive conversational tutors using a taxonomy derived from authentic ESL tutoring interactions.
Its \textit{student-move} taxonomy describes the interactional function of a learner and whether an attempted answer requires repair \citep{stolcke2000dialog,suresh2022talkmoves,lyster1997corrective}. Its \textit{tutor-strategy} taxonomy comprises 13 strategies spanning interaction management, instructional support, and language-focused feedback. It integrates general tutoring functions, such as prompting, backchanneling, and affective support, with instructional and language-focused moves, including topic-contingent teaching, error flagging, hinting, and direct correction~\citep{shute2008feedback,lyster1997corrective}. Together, the two representations operationalize two linked decisions: whether a learner turn warrants pedagogical intervention and, if so, which response strategy is appropriate.

Drawing on 260 authentic one-to-one English tutoring sessions conducted via text chat from the Teacher--Student Chatroom Corpus version 2 (TSCC v2)~\citep{caines2022tsccv2}, we construct \tactcorpus with 32,379 taxonomy annotations and a quality-controlled augmented training set. We then post-train Qwen3.5-4B through supervised fine-tuning and taxonomy-grounded Group Relative Policy Optimization (GRPO) to produce \tactutor.

On \tactbench, our strategy-balanced, optimization-aligned diagnostic benchmark comprising 78 authentic contexts, \tactutor scores 20.3 percentage points above its backbone and above all evaluated proprietary baselines under the same protocol. Beyond this targeted diagnostic, \tactutor matches or improves upon its backbone on all evaluated external educational benchmarks. In a blinded, within-participant study with 50 learners, \tactutor obtains the highest descriptive overall mean (5.54/7), outperforming its backbone across all four teaching-behavior dimensions and both frontier proprietary models overall. In summary, we contribute:


\begin{itemize}

\item An operational taxonomy for learner-responsive ESL tutoring, comprising 13 tutor strategies and a two-axis learner-behavior schema.
\item \tactcorpus, a model-assisted, human-audited resource derived from authentic tutoring dialogue with quality-controlled augmentations, and \tactbench, a strategy-balanced, optimization-aligned diagnostic benchmark on ESL tutoring.

\item \tactutor, an open 4B model post-trained through taxonomy-guided supervised fine-tuning and GRPO and evaluated through targeted diagnostics, external educational benchmarks, and blinded human interaction.

\end{itemize}

We release \tactcorpus, \tactbench, and the \tactutor model weights to support reproducible research on pedagogically grounded language tutoring.

%% file: sections/02-theoretical-framing.tex
\section{Taxonomy Design}
\input{tables/student-taxonomy}

This section defines the pedagogical decision space that connects learner-state annotation, tutor-strategy selection, and next-turn response generation. Existing taxonomies capture complementary but incomplete aspects of ESL tutoring. General tutoring schemes describe instructional and conversational functions but rarely indicate whether a learner's task response warrants repair \citep{graesser1995collaborative,graesser2005autotutor}. Corrective-feedback schemes characterize repair moves but often omit the learner turn's broader conversational role and the legitimate option of continuing without correction \citep{lyster1997corrective,hall200210}. We therefore develop an integrated taxonomy that represents two connected elements of ESL tutoring: the learner state to which the tutor responds and the instructional function of the next tutor turn; full related work is provided in the appendix section ``Additional Background and Related Work.''

The taxonomy has two linked components. The \emph{student-move taxonomy} records what the learner has done and the task-level status of that move. The \emph{tutor-strategy taxonomy} records the intended pedagogical function of the tutor's response. Together, these elements define the decision process in Figure~\ref{fig:tact-feedback}: given the instructional context and the student's latest move, select an appropriate tutor strategy and realize it as a natural-language response.

\input{Figures/fig-tact-feedback}

\subsection{Theoretical Basis and Construction}

The design follows the view that effective feedback is conditional: its timing, explicitness, and degree of elaboration should depend on the learner, task, and current response \citep{mason2001feedback,shute2008feedback}. We operationalize these factors for one-to-one ESL dialogue by representing learner proficiency and task difficulty as contextual metadata, the current response through student-move labels, and the type and amount of support through tutor-strategy labels. The labels expose the pedagogical decision for annotation and learning, while the target output remains a natural-language tutor response rather than a taxonomy label.

Expert annotators developed the coding scheme iteratively from authentic ESL dialogues. They first manually annotated 50 dialogue contexts while drafting student-move and tutor-strategy codebooks, reconciling overlapping categories, and refining definitions and examples through repeated coding. We then independently double-coded 20 held-out contexts for inter-annotator reliability before joint adjudication. We report observed agreement and Cohen's kappa~\citep{cohen1960coefficient}. Observed agreement is
\begin{equation}
P_o
=
\frac{1}{N}
\sum_{i=1}^{N}
\mathbf{1}\!\left[y_i^{(1)}=y_i^{(2)}\right].
\label{eq:observed-agreement}
\end{equation}
where \(y_i\) denotes the assigned label or label set. Cohen's kappa corrects this score for chance agreement:
\begin{equation}
\kappa
=
\frac{P_o-P_e}{1-P_e},
\quad
P_e
=
\sum_{c}p_c^{(1)}p_c^{(2)},
\label{eq:cohens-kappa}
\end{equation}
where \(P_e\) is estimated from annotator marginals. Reliability was computed separately for multi-label tutor strategy, student move type, and student move status. Tutor-strategy annotation reached \(P_o=0.760\) and \(\kappa=0.708\); student move type reached \(P_o=0.798\) and \(\kappa=0.722\); and student move status reached \(P_o=1.000\) and \(\kappa=1.000\). Disagreements informed the final codebook before corpus-scale annotation.

\subsection{Student-Move Representation}

The student-move taxonomy characterizes the learner's latest turn along two dimensions: \emph{move type} and \emph{move status} (Table~\ref{tab:student-taxonomy}). Move type distinguishes answers or task attempts, questions, statements, acknowledgments, and off-task or social interaction. Move status distinguishes an accepted answer, an answer requiring repair, and an open or non-evaluable move.

The two dimensions are constrained. Only \textit{Answer / Attempt} moves receive the status \textit{Adequate / Accepted} or \textit{Problematic / Needs Repair}; other move types receive \textit{Non-evaluable / Open}. Importantly, move status describes task-level evaluability rather than the grammatical well-formedness of the entire utterance. A learner's turn may contain a language issue but still function as an open question rather than an incorrect task attempt; distinguishing move type from status therefore determines whether the tutor should evaluate an answer, explain a point, or continue the interaction.

\subsection{Tutor Strategies and Decision Mapping}

The tutor-strategy taxonomy contains 13 strategies representing the intended pedagogical function of the next tutor turn (Table~\ref{tab:teacher-taxonomy}). It covers conversational and tutoring functions, such as prompting, backchanneling, and affective/social feedback, and also instructional feedback moves such as verification, correct response, topic-contingent teaching, response-contingent reformulation, error flagging, hinting, guided revision, clarification checking, and direct correction.

\input{tables/teacher-taxonomy}

Strategy labels describe what a response is intended to accomplish, not its surface wording. Different utterances may realize the same strategy, and multi-label annotation is allowed when one turn performs multiple functions, such as acknowledging the learner before providing a hint. The strategies also differ in assistance level: signaling an error preserves more learner agency than supplying a correction, whereas a targeted hint provides more support than a general invitation to try again.

The mapping from student moves to tutor strategies is conditional rather than deterministic. A problematic answer may call for an error signal, hint, guided revision, or direct correction depending on student proficiency, task difficulty, dialogue history, and prior support. Similarly, an accepted answer may be confirmed, elaborated, or followed by task progression. The taxonomy thus structures the central generation problem addressed by \tool: selecting an appropriate pedagogical action for the current student state and expressing that action as a coherent natural-language tutor response.

%% file: tables/student-taxonomy.tex
\begin{table*}[!t]
\centering
{\small
\setlength{\tabcolsep}{3pt}
\begin{tabular}{p{0.06\textwidth}p{0.22\textwidth}p{0.35\textwidth}p{0.33\textwidth}}
\toprule
\textbf{Axis} & \textbf{Label} & \textbf{Definition} & \textbf{Real dialogue example} \\
\midrule
Move type & Q -- Question / Inquiry & The learner asks for meaning, correctness, explanation, or usage. & ``What does `flying colours' mean?'' \\
Move type & A -- Answer / Attempt & The learner attempts the current exercise, sentence, translation, or target-language task. & ``The result of an exam was a relief for me." when the teacher asked the student to make a sentence using ``relief" or ``excitement". \\
Move type & S -- Statement / Explanation / Comment & The learner states an idea, explanation, example, or observation rather than directly answering. & ``I think group 1 is more emotional." when the teacher asked which preposition should be used with group 1 and group 2. \\
Move type & F -- Feedback / Acknowledgment & The learner acknowledges, accepts, rejects, or signals understanding of a teacher turn. & ``Exactly.'' after the teacher asks whether the exam is IELTS. \\
Move type & O -- Other / Social / Off-task & The learner gives social, emotional, filler, or non-task content. & ``Oh, It's difficult!'' \\
\midrule
Status & A -- Adequate / Accepted & Only valid when the move type is A; the learner's attempt satisfies the current requirement. & A correct fill-in answer accepted by the teacher. \\
Status & P -- Problematic / Needs Repair & Only valid when the move type is A; the learner's attempt is wrong, incomplete, unnatural, or needs revision. & ``carears advisor'' before the spelling correction ``careers.'' \\
Status & N -- Non-evaluable / Open & Fixed for Q, S, F, and O turns; also used when an answer-like turn cannot be judged as right or wrong. & ``In my last job I worked as a carears advisor...'' as open conversation context before the teacher targets spelling. \\
\bottomrule
\end{tabular}
}
\caption{Student-Move Taxonomy. Only Answer/Attempt turns can be Adequate or Problematic; other move types take N/Non-evaluable. Examples are from TSCC-derived contexts.}
\label{tab:student-taxonomy}
\end{table*}

%% file: Figures/fig-tact-feedback.tex
\begin{figure}[t]
\setlength{\fboxsep}{6pt}

\includegraphics[width=\columnwidth]{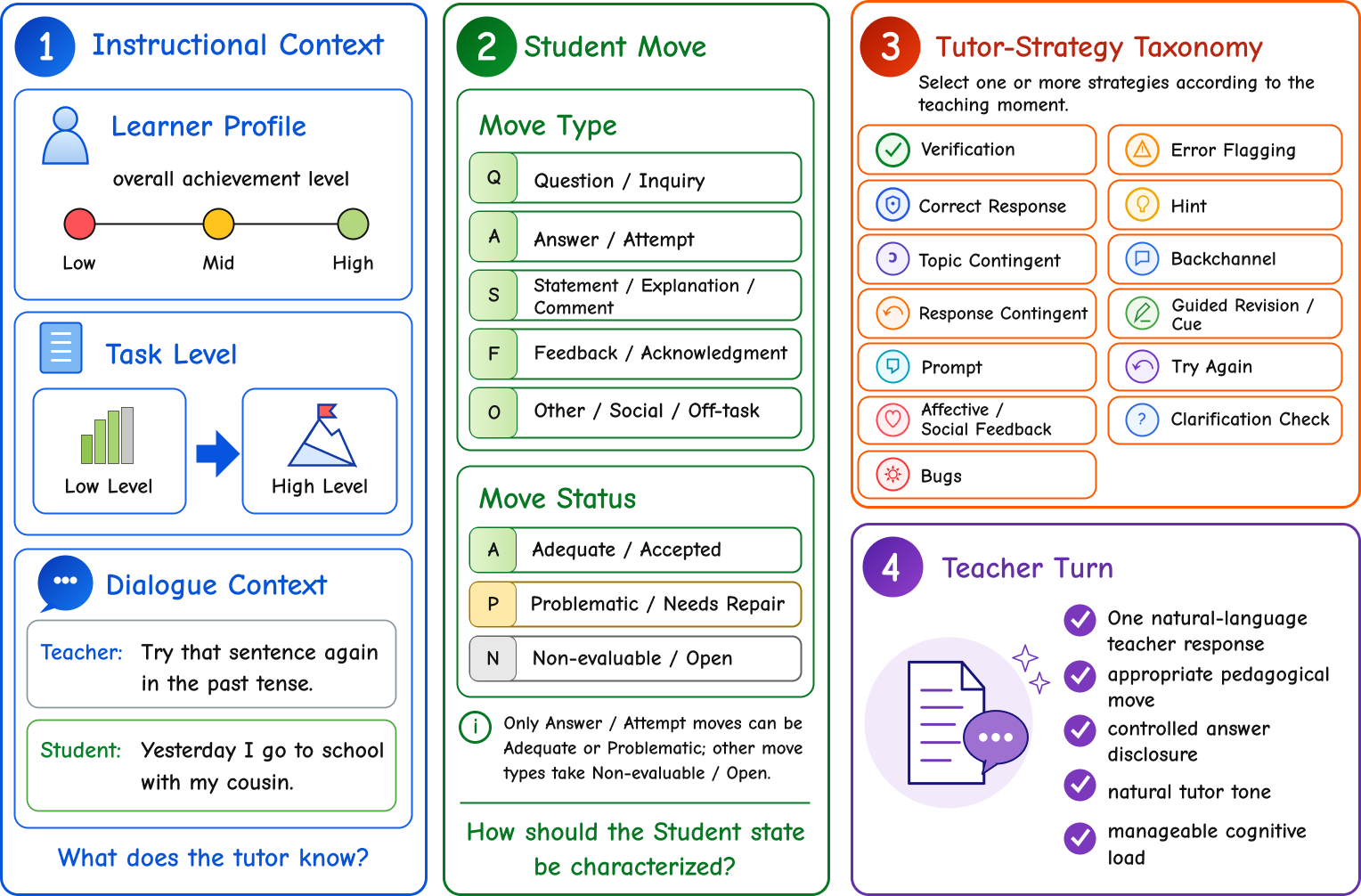}
\caption{\tool feedback-decision framing for next-turn ESL tutoring. The tutor conditions on the instructional context, characterizes the current student move, selects one or more tutor strategies, and realizes them as a natural teacher turn with appropriate pedagogy, controlled answer disclosure, natural tone, and manageable cognitive load.}
\label{fig:tact-feedback}
\end{figure}

%% file: tables/teacher-taxonomy.tex
\begin{table*}[t]
\centering
{\small
\setlength{\tabcolsep}{2pt}
\begin{tabular}{p{0.15\textwidth}p{0.42\textwidth}p{0.4\textwidth}}
\toprule
\textbf{Tutor label} & \textbf{One-sentence definition} & \textbf{Real held-out/context example} \\
\midrule
Verification & Confirms whether an answer is right or wrong without extra teaching content. & ``That's right!'' after the learner provides a correct answer. \\
Correct Response & Gives the specific correct answer without extra explanation. & ``It's careers!'' after the learner misspells the word as ``carears''. \\
Topic Contingent & Adds relevant teaching content when the learner turn is not simply correct or incorrect. & ``In this context, it means you passed with a high standard; you got a high mark basically.'' after the student asks what "flying colours" means. \\
Response-Contingent & Builds on a correct answer with a more natural or more complete version. & A teacher accepts the student's wording, then offers a more idiomatic phrasing. \\
Bugs & Diagnoses and explains a specific student error or misconception. & ``The noun \textit{confusion} is commonly used with verbs like \textit{cause}; for traffic, we would say something like \textit{chaos}.'' after the student produces ``The accident made a big confusion on the traffic.'' \\
Error Flagging & Points out that a particular part is wrong or suspicious without fully correcting it. & ``Definitely not \textit{the unemployment}.'' after the student produces ``The unemployment is high in the whole country.'' \\
Hint & Gives a directional clue while leaving the learner to revise. & A cue to attend to word order, verb form, preposition, or intended meaning. \\
Guided revision / Cue & Provides a frame, blank, option set, contrast, or judgment task that constrains revision. & ``It is very \_\_\_\_\_\_\_ that you will get a double six.'' \\
Try Again & Explicitly asks the learner to try again. & ``Not quite--try that sentence again.'' \\
Clarification Check & Checks the intended meaning of an unclear or possibly inaccurate learner expression. & ``Do you mean, `I was impressed by the film'?'' after the student produces ``The film gave me impression.'' \\
\midrule
Prompt & Asks for the next answer, example, sentence, or task step. & ``Can I ask you to write a sentence using the following information: Pacific/biggest?'' \\
Backchannel & Briefly acknowledges the learner without substantive correction or teaching. & ``OK, thanks'' after the student attempts an answer. \\
Affective/ Social Feedback & Provides encouragement, praise, humor, greeting, or emotional support. & ``I thought you seemed very fluent...'' before continuing the conversation. \\
\bottomrule
\end{tabular}
}
\caption{Tutor-Strategy Taxonomy. Labels cover conversational/tutoring functions and instructional feedback moves. Examples are drawn from TSCC-derived contexts.}
\label{tab:teacher-taxonomy}
\end{table*}

%% file: sections/03-taxonomy-and-data-construction.tex
\section{\tactcorpus: Taxonomy-Guided Data Construction}

We derive \tactcorpus from TSCC v2, a corpus containing 260 authentic one-to-one English lessons conducted through private online chatrooms, involving 2 teachers and 13 learners across CEFR levels B1--C2 \citep{caines2022tsccv2}. We merge consecutive messages from the same participant while preserving the original content and lesson order, then convert each eligible merged teacher turn into one next-teacher-turn response-generation instance. The visible input contains the preceding dialogue, the latest student turn, the learner's overall achievement level, and the task level, and the prediction target is the original next teacher turn.
For training instances, available student-move labels may additionally be included as structured learner-state metadata. For diagnostic evaluation, these labels are excluded from the policy-model input because manually annotated student states would not normally be available during real-world deployment. The model must instead infer the learner state from the dialogue itself.
The original teacher response, tutor-strategy labels, student-move annotations, and rubric metadata are retained as hidden fields. These fields are not shown to the policy model during test-time generation. They are used as supervised targets, reward anchors, and diagnostic labels, as detailed in the appendix section ``Prompt Protocol.''

\subsection{Dataset Statistics and Splits}

\tactcorpus contains 32,379 annotation records, including tutor-strategy labels, student-move labels and augmentation metadata. These records are associated with the normalized dialogue turns and provide the structured supervision needed for taxonomy-guided training and evaluation. After filtering and converting dialogues into next-teacher-turn response-generation instances, the training split contains 2,702 examples. \tactbench contains 78 manually verified tutoring contexts, selected to support controlled comparison across different pedagogical strategies and learner states. A summary of \tactcorpus and \tactbench, including dataset statistics and model configuration, is provided in Table~\ref{tab:revised-dataset-stats}.

%% file: sections/04-post-training-method.tex
\input{Figures/fig-tact-pipeline}

\section{Taxonomy-Guided Post-Training and Evaluation}
\label{sec:training-evaluation}

We use the student-move and tutor-strategy taxonomies to connect \tactutor post-training and \tactbench diagnostic evaluation within a shared pedagogical decision space. As illustrated in Figure~\ref{fig:tact-pipeline}, our pipeline consists of three phases: (1) supervised fine-tuning (SFT) teaches \tactutor how appropriate teacher responses are expressed in authentic tutoring contexts; (2) taxonomy-grounded reward optimization encourages \tactutor to select an appropriate instructional strategy and level of support; and (3) \tactbench measures whether \tactutor can make these decisions using only information that would be available during deployment.

\subsection{Phase I: Supervised Fine-Tuning}
\label{sec:sft}

We first perform parameter-efficient SFT with LoRA \citep{hu2021lora} on response-generation instances constructed from annotated TSCC v2 dialogues. Each instance is formatted as a chat transcript, where tutoring instructions, learner profile, and previous dialogue turns form the context, and the original next teacher turn serves as the generation target.

To explicitly model learner states, half of the SFT instances additionally include student-move hints in the system message. These hints provide the type and status of the learner's latest move, while the remaining instances preserve the original dialogue-only setting. Tutor-strategy labels are used as semantic annotations rather than generation targets: they describe the intended pedagogical function of each response, while student-move labels characterize the learner state that motivates this function.

\input{tables/revised-dataset-stats}

We additionally construct a best-of-\(n\) aligned SFT variant, where a pedagogical judge selects high-quality candidate responses that are acceptable and consistent with the intended tutoring strategy. This reduces dependence on a single reference response and improves supervision diversity. 
Details of candidate generation and filtering are provided in the appendix sections ``Prompt Protocol'' and ``Prompt Listings.''

\input{tables/tactbench-rubric}

\subsection{Phase II: Group Relative Policy Optimization}
\label{sec:grpo}

Although SFT teaches \tactutor to imitate teacher responses, it does not directly optimize whether a generated response selects an appropriate instructional action. We therefore further optimize \tactutor using GRPO \citep{shao2024deepseekmath}.

For each dialogue context, the policy generates multiple candidate responses. The policy only receives deployment-available information and does not access reference responses, gold tutor-strategy labels, or evaluation rubrics. A frozen pedagogical judge evaluates each candidate using the hidden reference responses, gold taxonomy labels with their codebook definitions, and the evaluation rubric, producing five quality scores (Table~\ref{tab:tactbench-rubric}) and three binary failure indicators for acceptability, leakage/over-helping, and off-task errors.

For each dimension \(d\), the raw score \(q_d\in\{1,\ldots,5\}\) is normalized as \(S_d=(q_d-1)/4\). We define \(r_{\mathrm{task}}\) as the mean normalized score and combine it with grounding, format, and strategy rewards:

\begin{equation}
\begin{aligned}
r_{\mathrm{base}} ={}& r_{\mathrm{task}}
+ \alpha r_{\mathrm{ground}}
+ \beta r_{\mathrm{format}} \\
&+ \gamma r_{\mathrm{strategy}}
+ r_{\mathrm{penalty}} .
\end{aligned}
\end{equation}

The grounding term \(r_{\mathrm{ground}}\) rewards cues consistent with the intended instructional function. The format term \(r_{\mathrm{format}}\) ensures that responses are usable teacher turns by penalizing malformed or undesirable outputs. The strategy term \(r_{\mathrm{strategy}}\) captures strategy-specific behaviors, such as learner-centered support for hints and guided revision, explicit error signaling, and concise verification or affective responses. Together, these components make the reward transparent and auditable rather than an opaque optimization of a single benchmark score.

The penalty term uses judge-level guardrails:

\begin{equation}
r_{\mathrm{penalty}}
=
\delta a-\lambda l-\mu o .
\end{equation}

Here, \(a\) indicates an acceptable response, \(l\) indicates answer leakage or over-helping, and \(o\) indicates an off-task or incoherent response. We set \(\alpha=0.2\), \(\beta=0.1\), \(\gamma=0.3\), \(\delta=0.2\), \(\lambda=\mu=1.0\), and clip rewards to \([-3,2]\).

The final GRPO objective additionally applies a KL constraint against the frozen SFT reference policy, following standard policy optimization practice. With \(\eta_{\mathrm{KL}}=0.02\), this constraint limits excessive policy drift while allowing optimization toward improved pedagogical decisions.

Reward component definitions and implementation details are provided in the appendix section ``Training Configurations.''

\subsection{Phase III: Diagnostic Evaluation}
\label{sec:diagnostic-evaluation}

We evaluate \tactutor under deployment-like conditions using \tactbench. During generation, the model receives only the dialogue history, latest learner turn, learner achievement level, task level, and system prompt. It does not access student-move labels, gold tutor-strategy labels, or the original teacher response. The model must infer the learner state and produce an appropriate next teacher turn from observable context.

After generation, the diagnostic judge evaluates each response using the hidden reference response, gold taxonomy labels with their codebook definitions, and the evaluation rubric. The rubric measures five dimensions reflecting strategy selection, contextual responsiveness, pedagogical support, learner-level suitability, and leakage control (Table~\ref{tab:tactbench-rubric}).

Each \tactbench item evaluates open-ended next-teacher-turn generation rather than label prediction: the model is scored by the response it produces, while the reference response, gold taxonomy labels, and codebook definitions remain hidden from the policy and are used only by the judge. We report the mean normalized score across the five dimensions as \(\mathrm{TACT\_Overall}\), together with three auxiliary failure rates: accept rate, leak-or-overhelp rate, and off-task-or-incoherent rate.

These metrics separate different tutoring failure modes. A fluent response may still fail by selecting an inappropriate teaching move, providing excessive assistance, or mismatching the learner's need. Therefore, \tactbench evaluates not only response quality but also the pedagogical decision process underlying the generated response.

Prompt protocols, training configurations, computing infrastructure, and randomness-control details are provided in the appendix sections ``Training Configurations,'' ``Prompt Protocol,'' and ``Prompt Listings.''

%% file: Figures/fig-tact-pipeline.tex
\begin{figure*}[t]
\centering
\includegraphics[width=0.9\textwidth]{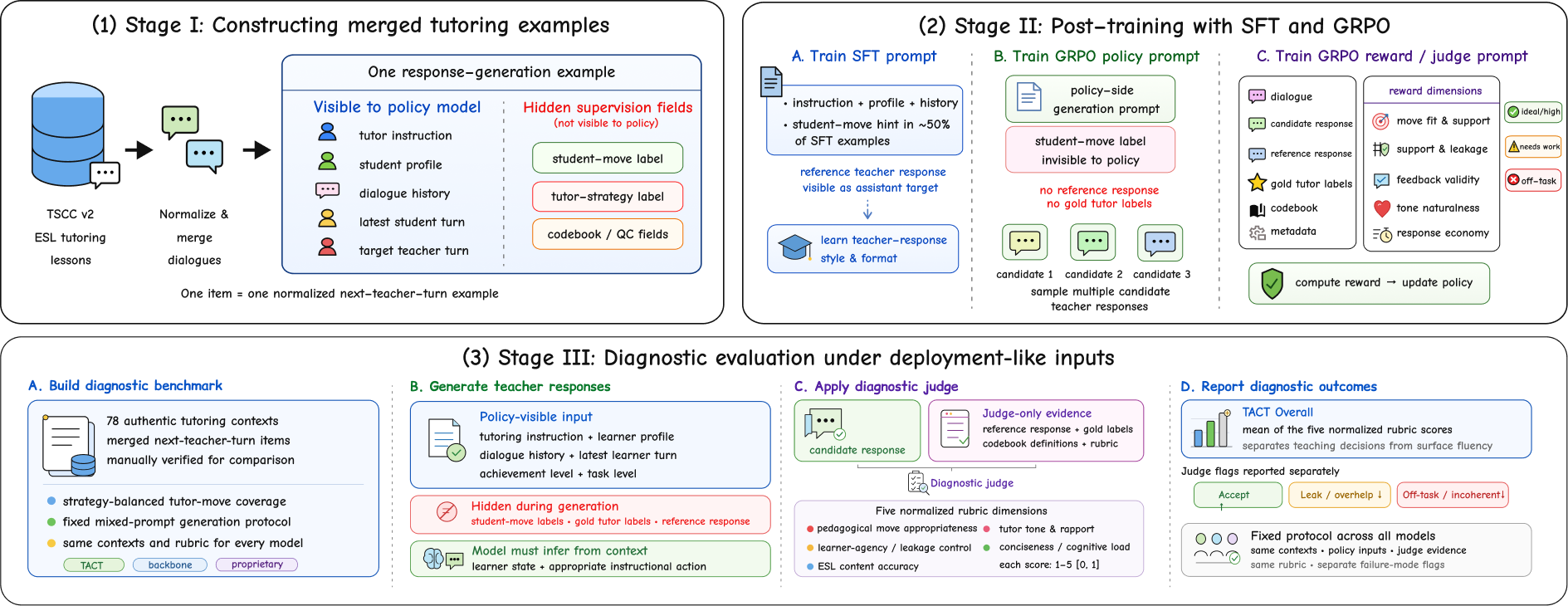}
\caption{\tool pipeline: Stage I builds \tactcorpus from ESL dialogues; Stage II trains \tactutor with SFT and taxonomy-guided GRPO; Stage III evaluates next-turn responses with \tactbench under deployment-like inputs.}
\label{fig:tact-pipeline}
\end{figure*}

%% file: tables/revised-dataset-stats.tex
\begin{table}[h]
\centering
{\small
\setlength{\tabcolsep}{3pt}
\begin{tabular}{lr}
\toprule
\textbf{Quantity} & \textbf{Value} \\
\midrule
TSCC v2 lessons & 260 \\
Total annotation records & 32,379 \\
Tutor strategy labels & 13 \\
Student-move axes & 2 \\
Standard training examples & 2,702 \\
Verified diagnostic contexts & 78 \\
Backbone model & Qwen3.5-4B \\
\bottomrule
\end{tabular}
}
\caption{\tactcorpus and \tactbench summary. Annotation count follows the project release summary; example counts are from the current training and diagnostic splits.}
\label{tab:revised-dataset-stats}
\end{table}

%% file: tables/tactbench-rubric.tex
\begin{table}[b]
\centering
{\small
\setlength{\tabcolsep}{3pt}
\begin{tabular}{@{}p{0.1\columnwidth}p{0.22\columnwidth}p{0.6\columnwidth}@{}}
\midrule
Move & Pedagogical move & Selects a teacher action matching the gold tutor-strategy function and local dialogue need. \\
Agency & Learner agency & Gives appropriate support without doing the learner's work unless direct correction is warranted. \\
ESL & Content accuracy & Gives correct English-language feedback, explanation, examples, or correction. \\
Tone & Tutor rapport & Sounds natural, supportive, and locally appropriate as a teacher turn. \\
Conc. & Cognitive load & Stays focused and avoids unnecessary length, distraction, or too many teaching points. \\
\bottomrule
\end{tabular}
}
\caption{Five \tactbench rubric dimensions. Each is scored from 1 to 5 and normalized to $[0,1]$; 
}
\label{tab:tactbench-rubric}
\end{table}

%% file: sections/05-evaluation-and-results.tex
\section{Evaluation and Results}
In this section, we report our evaluation pipeline and results.
\subsection{Diagnostic Results}

\input{tables/main-results}

Table~\ref{tab:main-results} summarizes the \tactbench diagnostic results. It reports the five normalized rubric dimensions and \(\mathrm{TACT\_Overall}\), while the auxiliary Accept, Leak, and Off-task judge flags are reported separately in Appendix Table~\ref{tab:tactbench-judge-flags}. \tactutor first uses SFT to adapt the Qwen3.5-4B backbone to authentic ESL tutor-turn format, then uses taxonomy-grounded GRPO because reference imitation alone cannot distinguish a merely similar teacher turn from one that makes a defensible tutoring decision. This post-training raises \(\mathrm{TACT\_Overall}\) from 0.629 for the backbone to 0.832; the auxiliary flags also improve, with accept rate increasing from 0.603 to 0.872 and leak-or-overhelp falling from 0.346 to 0.026. Among strong leaderboard baselines, qwen3.7-max is closest in Overall score (0.829), GLM-5.2 leads on Move and Tone, and GPT-5.5 leads on ESL; nevertheless, \tactutor achieves the highest Overall score despite using an open 4B backbone.

\subsection{Human Evaluation}
\input{tables/human-eval}

We conducted a blinded, within-participant interactive evaluation with 50 real-world learners via Prolific crowdsourcing~\cite{prolific2024prolific}. Each participant completed four short English-learning conversations—one with each of four anonymous tutors (Tutor~A--D)—during an approximately 40-minute online study and received US\$10 in compensation, yielding 200 conversations in total. Each conversation contained at least five learner turns; model identity was hidden, and tutor order and topic assignment were counterbalanced. After each conversation, participants rated the tutor from 1 to 7 on \textit{guiding}, \textit{scaffolding}, \textit{encouragement}, and \textit{support for learner self-correction}.

As shown in Table~\ref{tab:human-eval}, \tactutor achieved the \textbf{highest mean score} (5.54), within 0.09 points of both proprietary tutors, and was rated above its backbone across all four dimensions ($+0.39$ overall). The gains were not uniform, but concentrated in \textit{encouragement} ($+0.58$), \textit{guiding} ($+0.40$), and \textit{scaffolding} ($+0.38$). This pattern suggests that taxonomy-aligned post-training changes the tutor’s interactional profile toward more supportive and process-oriented responses, rather than producing only a general improvement in conversational quality. \textit{Self-correction} improved more modestly ($+0.20$), plausibly because it depends not only on the tutor’s response but also on whether the learner subsequently recognizes and repairs the error. Overall, the results show that the learned behaviors are perceptible to learners during interaction, while motivating trajectory-level training and evaluation centered on learner uptake.

\subsection{External Benchmarks}
Appendix Table~\ref{tab:external-benchmarks} reports transfer checks on external educational benchmarks covering math tutoring, STEM tutoring, long-history personalization, and interactive tutor simulation. \tactutor shows no degradation relative to the corresponding Qwen baselines: MRBench~\citep{maurya2025mrbench} macro DAMR rises from 0.743 to 0.880, TutorBench~\citep{srinivasa2025tutorbench} weighted score from 0.497 to 0.519, LongTutor~\citep{li2026longtutor} teaching quality from 3.712 to 3.834, and the DeepTutor~\citep{zhao2026deeptutor} smoke score from 3.565 to 3.627. These results suggest that our method improves \tactutor on in-domain ESL tutoring while preserving broader tutoring capability.

%% file: tables/main-results.tex
\begin{table}[!t]
\centering
{\small
\setlength{\tabcolsep}{2pt}
\begin{tabular}{@{}>{\raggedright\arraybackslash}p{0.28\columnwidth}*{5}{>{\centering\arraybackslash}p{0.09\columnwidth}}>{\centering\arraybackslash}p{0.13\columnwidth}@{}}
\toprule
\textbf{Model} & \textbf{Move} & \textbf{Agency} & \textbf{ ESL} & \textbf{Tone} & \textbf{Conc.} & \shortstack{\textbf{Overall}}\\
\midrule
\tactutor (ours) & 0.692 & \underline{\textbf{0.801}} & 0.917 & 0.872 & \underline{\textbf{0.878}} & \underline{\textbf{0.832}} \\
\midrule
qwen3.7-max & 0.744 & 0.750 & 0.962 & 0.926 & 0.766 & 0.829 \\
GLM-5.2 & \textbf{0.756} & 0.708 & 0.949 & \textbf{0.939} & 0.734 & 0.817 \\
Claude-Opus-4.6 & 0.734 & 0.734 & 0.933 & 0.904 & 0.728 & 0.806 \\
Kimi-k2.6 & 0.696 & 0.724 & 0.936 & 0.913 & 0.724 & 0.799 \\
Gemini-2.5-Pro & 0.676 & 0.728 & 0.907 & 0.872 & 0.808 & 0.798 \\
DeepSeek-V4-Pro & 0.647 & 0.686 & 0.917 & 0.862 & 0.798 & 0.782 \\
GPT-5.5 & 0.724 & 0.619 & \textbf{0.974} & 0.843 & 0.712 & 0.774 \\
Kimi-k2.5 & 0.628 & 0.676 & 0.891 & 0.827 & 0.763 & 0.757 \\
Qwen-max & 0.526 & 0.538 & 0.888 & 0.821 & 0.718 & 0.698 \\
Qwen3.5-4B & 0.516 & 0.516 & 0.798 & 0.756 & 0.561 & 0.629 \\
Kimi-k3 & 0.522 & 0.542 & 0.779 & 0.615 & 0.631 & 0.618 \\
\bottomrule
\end{tabular}
}
\caption{\tactbench diagnostic results on 78 ESL tutoring contexts. All models use the same DeepSeek-based judge and mixed-prompt protocol. Auxiliary judge flags are reported in Appendix Table~\ref{tab:tactbench-judge-flags}. Bold marks the best value per column; \underline{underline} marks \tactutor{}'s core pedagogical metrics.}
\label{tab:main-results}
\end{table}

%% file: tables/human-eval.tex
\begin{table}[b]
\centering
{\small
\begin{tabular}{lccccc}
\toprule
\textbf{Model} & \textbf{Guid.} & \textbf{Scaf.} & \textbf{Enc.} & \textbf{S-corr.} & \textbf{Mean} \\
\midrule
\tactutor (ours) & 5.56 & \textbf{5.54} & \textbf{5.76} & 5.30 & \textbf{5.54} \\
DeepSeek-V4-Pro & \textbf{5.60} & 5.36 & 5.52 & 5.38 & 5.47 \\
GLM-5.2 & 5.50 & 5.36 & 5.46 & \textbf{5.48} & 5.45 \\
Qwen3.5-4B base & 5.16 & 5.16 & 5.18 & 5.10 & 5.15 \\
\bottomrule
\end{tabular}
}
\caption{Blinded human evaluation (Prolific, $n=50$). Mean learner ratings use a 1--7 scale; bold marks the best value per column.}
\label{tab:human-eval}
\end{table}

%% file: sections/06-discussion.tex
\section{Discussion}

\tool bridges tutoring theory and model post-training by shifting the objective from imitating the next teacher utterance to selecting an appropriate feedback action for the learner's current state and realizing it naturally. This distinction is important because authentic teacher data contain both pedagogical signal and conversational noise: direct imitation may reproduce brevity, acknowledgments, or topic management without learning the underlying instructional decision. The taxonomy exposes that decision, with SFT providing domain style and GRPO optimizing the intended tutoring function.

Accordingly, the reference response is a pedagogical anchor rather than a unique target. Multiple responses may be valid if they address the learner's immediate need, preserve agency, and maintain a coherent lesson trajectory. Evaluation and reward should therefore recognize defensible tutoring functions rather than require mechanical agreement with the reference wording or label.

More broadly, \tactbench provides an auditable training signal, diagnostic tool, and standardized protocol for tutoring decisions. In-domain results measure alignment with the \tool construct, while external benchmarks serve as transfer sanity checks. Its decomposed scores also make failures actionable: errors in strategy selection, answer disclosure, ESL accuracy, tone, conciseness, or formatting point respectively to improvements in taxonomy coverage, reward design, prompting, data filtering, or model capacity.

%% file: sections/07-conclusion-and-limitations.tex
\section{Conclusion and Limitations}

We presented \tool, a taxonomy-aligned framework for post-training and evaluating ESL tutor response generators. Starting from TSCC v2, \tool constructs \tactcorpus, trains Qwen3.5-4B into \tactutor with SFT and GRPO, and evaluates candidate teacher turns with the five-dimensional \tactbench rubric.
The main limitation is that next-turn response quality is not the same as long-term learning. A response can be locally well scaffolded while its effect depends on learner motivation, prior knowledge, repeated interaction, and classroom context. A second limitation is label abstraction: a compact taxonomy is useful for training and diagnosis, but real teachers blend instruction, affect, pacing, and rapport in ways that are not always separable. A third limitation is evaluation coupling. Because \tactbench is designed around a specified \tool construct, improvements should be interpreted alongside disaggregated rubric dimensions, judge flags, and external transfer checks rather than as universal proof of better teaching.
Future work should measure downstream learner behavior, such as whether subsequent learner turns successfully self-repair. \tactcorpus and \tactbench are intended to make those audits possible.


%% file: sections/07-ethical-statement.tex


%% file: sections/08-appendix.tex
\section{Appendix}
\label{app:appendix}

\subsection{Additional Background and Related Work}
\label{sec:additional-background-related-work}

\paragraph{Prompted and language-learning tutors.}
Current LLM tutors commonly introduce pedagogical control at inference time. Teaching with AI, Khanmigo-style deployments, Study Mode, and Claude for Education use prompting or product-level interaction policies to encourage guided questioning, hints, and Socratic dialogue \citep{openai2023teaching,openai2023khan,openai2025study,anthropic2025education}. Language-learning systems add proficiency and language-choice constraints: CEFR-prompted tutors condition responses on proficiency but may drift from the requested level over extended interaction, while pedagogical code-switching systems explicitly control when a learner's first language should support comprehension \citep{almasi2025alignment,liu2024codeswitching}. Two systems are particularly close to conversational tutoring. \citet{vanzo2025gpt} use prompting to turn GPT-4 into an interactive homework tutor, whereas BIPED first predicts a tutor act and then conditions an LLM on that act to generate an ESL response \citep{kwon2024biped}. Together, these systems show the value of explicit pedagogical control, but locate that control primarily in prompts, proprietary orchestration, or a separate planning stage.

\paragraph{Feedback and dialogue-action background.}
The action space for a conversational language tutor draws on several research traditions. Formative-feedback work distinguishes feedback by timing, specificity, elaboration, and the amount of information supplied \citep{shute2008feedback,mason2001feedback}. Second-language corrective-feedback research distinguishes recasts, explicit correction, clarification requests, metalinguistic feedback, elicitation, and repetition, which vary in directness and in the opportunity left for learner self-repair \citep{lyster1997corrective}. Dialogue-act and classroom-talk taxonomies contribute complementary interactional functions such as answering, acknowledging, prompting, and maintaining participation \citep{stolcke2000dialog,suresh2022talkmoves}. These traditions explain why conversational language tutoring must model both the learner's dialogue action and the teacher's decision about whether and how strongly to intervene.

\paragraph{External tutor benchmarks and systems.}
MRBench introduces an eight-dimensional taxonomy for evaluating mathematical tutor responses and releases 192 conversations with 1,596 human-annotated responses from LLM and human tutors \citep{maurya2025mrbench}. TutorBench evaluates 1,490 expert-curated high-school and AP items spanning adaptive explanation, actionable feedback, and hint generation; each item is paired with a sample-specific rubric for LLM-based judging \citep{srinivasa2025tutorbench}. Both benchmarks focus on whether a response is useful as teaching rather than merely correct, but primarily evaluate isolated short-form tutoring situations.

DeepTutor moves from response-only tutoring toward an open-source agentic learning workspace. It combines citation-grounded problem tutoring, difficulty-calibrated question generation, and dynamic learner memory, and evaluates personalized interaction with a profile-driven student simulator \citep{zhao2026deeptutor}. LongTutor targets a complementary limitation of short-context evaluation: it evaluates long-term personalized tutoring through three progressive tasks---historical evidence acquisition, knowledge-state diagnosis, and adaptive teaching action---using expert-annotated learning histories \citep{li2026longtutor}. These systems broaden the evaluation target from a single response to personalization, memory, and longer-term learner state. In our experiments, MRBench~\citep{maurya2025mrbench}, TutorBench~\citep{srinivasa2025tutorbench}, DeepTutor~\citep{zhao2026deeptutor}, and LongTutor~\citep{li2026longtutor} are therefore used as complementary transfer checks rather than as directly comparable substitutes for the ESL-specific diagnostic benchmark.

\section*{Ethical Statement}

This work uses anonymized ESL tutoring data from TSCC v2 and reports aggregate model and learner-evaluation results. The human evaluation recruited participants through Prolific, compensated them for the study, and hid model identities during rating. Released data and evaluation resources should preserve anonymization and avoid exposing private learner or tutor information.

\section*{Declaration of Generative AI and AI-Assisted Technologies in the Writing Process}

During the preparation of the \tool manuscript, the authors used ChatGPT only to improve readability and language. The authors reviewed, edited, and verified all AI-assisted language revisions, including the descriptions of the \tool framework, \tactcorpus, \tactbench, and \tactutor, and take full responsibility for the content of this work.


\subsection{Supporting Figures}
\label{app:supporting-figures}

Appendix Figure~\ref{fig:mason-feedback-framework} provides the feedback-decision background used to motivate the taxonomy variables.

Appendix Figure~\ref{fig:data_segmentation} summarizes how the raw tutoring logs are converted into model-ready training and evaluation instances.

Appendix Figure~\ref{fig:tscc-example-lesson} shows the chatroom format of the authentic TSCC v2 English tutoring lessons used as source data.

Appendix Figures~\ref{fig:human-eval-entry} and~\ref{fig:human-eval-chat} illustrate the human-evaluation platform used in our study, showing the entry and consent page and the anonymous tutor conversation interface, respectively.

\input{Figures/fig-mason-feedback-framework}
\input{Figures/Data_Segmentation}
\input{Figures/fig-tscc-example-lesson}
\input{Figures/human-eval-entry}
\input{Figures/human-eval-chat}

\subsection{Supporting Tables}
\label{app:supporting-tables}

\subsubsection{Training Configurations}
\label{app:training-configs}

Appendix Table~\ref{tab:sft-config} lists the supervised fine-tuning setup used to adapt the base model to ESL tutor-response generation.

\input{tables/SFT_Config}

Appendix Table~\ref{tab:grpo-config} lists the GRPO setup used after SFT, including the reward source, rollout settings, and regularization-relevant choices.

\input{tables/GRPO_config}

Appendix Table~\ref{tab:external-benchmarks} lists TACTutor performance on external benchmarks.

\input{tables/external-benchmarks}

\paragraph{Reward design and transfer preservation.}
The GRPO reward was designed to improve taxonomy-aligned ESL tutoring behavior while preserving general tutoring capabilities. Beyond task-specific pedagogical quality assessment, the reward incorporates complementary signals, including grounding, response-format constraints, strategy-shaping terms, and penalties for answer leakage and off-task behavior. These components target broader tutoring abilities such as providing actionable guidance, maintaining coherent multi-turn interactions, preserving learner agency, and producing concise, appropriately toned responses. These abilities are also important evaluation dimensions in external tutoring benchmarks: guidance and actionability are emphasized in MRBench and TutorBench, while coherence, personalization, history use, and multi-turn teaching ability are central to LongTutor and DeepTutor. Thus, the reward was designed not only for the target ESL tutoring objective, but also to avoid over-specialization toward a narrow evaluation criterion. The preserved performance on external benchmarks in Appendix Table~\ref{tab:external-benchmarks} is consistent with this transfer-preserving design.

\paragraph{Hyperparameter development and run counts.}
During development, we varied the SFT checkpoint used for downstream initialization, the GRPO checkpoint selected for evaluation, and the generation hyperparameters used during GRPO rollouts. The final SFT initialization and GRPO checkpoint were selected based on checkpoint-level model selection results under the standard development protocol, with auxiliary quality indicators used as additional selection criteria. Unless otherwise stated, each automated result reported in the diagnostic and guardrail tables is computed from one standard-protocol evaluation run with one generated response per model and item at temperature 0, followed by one judge pass. Reported automated scores are therefore single-run scores rather than averages over multiple random seeds.

\paragraph{Randomness control.}
For training, we used the Hugging Face/TRL trainer seed mechanism with seed 42. This seed controls the trainer-level Python, NumPy, and PyTorch random number generators used for adapter initialization or loading, data ordering, and stochastic training operations. No separate data seed was specified. The SFT run used deterministic data order with dataset shuffling disabled, while the GRPO run used trainer-seeded
dataset shuffling. GRPO rollouts were stochastic, using eight samples per prompt with temperature 0.9 and top-\(p=0.95\). The mixed prompt assignment used a fixed prompt-assignment seed, 20260711, encoded in the prompt variant \texttt{mixed\_1to1to1\_seed20260711}. For standard evaluation, candidate generation used temperature 0 and max length 512, and all reported scores were computed from the stored generations and judge outputs. Because GPU kernels and the external LLM judge may not be bitwise deterministic, we retained the generated responses, judgments, run manifests, and score files used to compute the reported numbers.

Experiments were run on a shared 8-GPU server. Unless otherwise noted, each SFT or GRPO training process used one CUDA-visible NVIDIA A800 80GB GPU. Appendix Table~\ref{tab:compute-config} summarizes the computing infrastructure and software environment.

\input{tables/Compute_Config}

\subsubsection{Prompt Protocol}
\label{app:prompt-protocol}

Appendix Table~\ref{tab:prompt-protocol} summarizes the prompt protocol and execution branches used across SFT, GRPO, and diagnostic evaluation. It identifies what information is visible to policy-side generation calls and judge-side scoring calls at each stage.

\input{tables/prompt-protocol}

Complete prompt skeletons are listed separately in the appendix section ``Prompt Listings.''

\subsubsection{Evaluation Tables}



Appendix Table~\ref{tab:tactbench-judge-flags} reports auxiliary diagnostic guardrails for the same \tactbench runs summarized in the main text.

\input{tables/tactbench-judge-flags}

Appendix Table~\ref{tab:tactbench-variation} reports per-item variation and bootstrap confidence intervals for the core diagnostic runs, complementing the aggregate means in the main results table.

\input{tables/tactbench-variation}

Appendix Table~\ref{tab:tactbench-test-cases} provides representative judged cases to illustrate how the rubric distinguishes fluent but pedagogically mismatched responses from appropriate ones.

\input{tables/tactbench-test-cases}

\subsection{\tactbench Judge Setting}
\label{app:tactbench-judge-setting}

For \tactbench diagnostic evaluation, the judge uses the DeepSeek-V4-Flash API model with the diagnostic judge prompt shown in Appendix Listing~\ref{lst:tactbench-judge-prompt}. The judge request sets \texttt{temperature=0.0}, \texttt{max\_tokens=2048}, and uses \texttt{max\_tokens} as the token-limit parameter. Judge thinking is disabled with \texttt{thinking=\{type: disabled\}} and \texttt{chat\_template\_kwargs=\{enable\_thinking: false\}}. We use the DeepSeek endpoint \url{https://api.deepseek.com}, a request timeout of 180 seconds, two retries, and a retry sleep of 2.0 seconds. No \texttt{top\_p}, \texttt{top\_k}, presence-penalty, or frequency-penalty parameter is set in the judge request.

\FloatBarrier

\subsection{Prompt Listings}
\label{app:prompt-listings}

The following listings provide compact prompt skeletons corresponding to the protocol tables above.

\input{sections/ppx-restore}

%% file: Figures/fig-mason-feedback-framework.tex
\begin{figure*}[!tbp]
\centering
\includegraphics[width=0.78\textwidth]{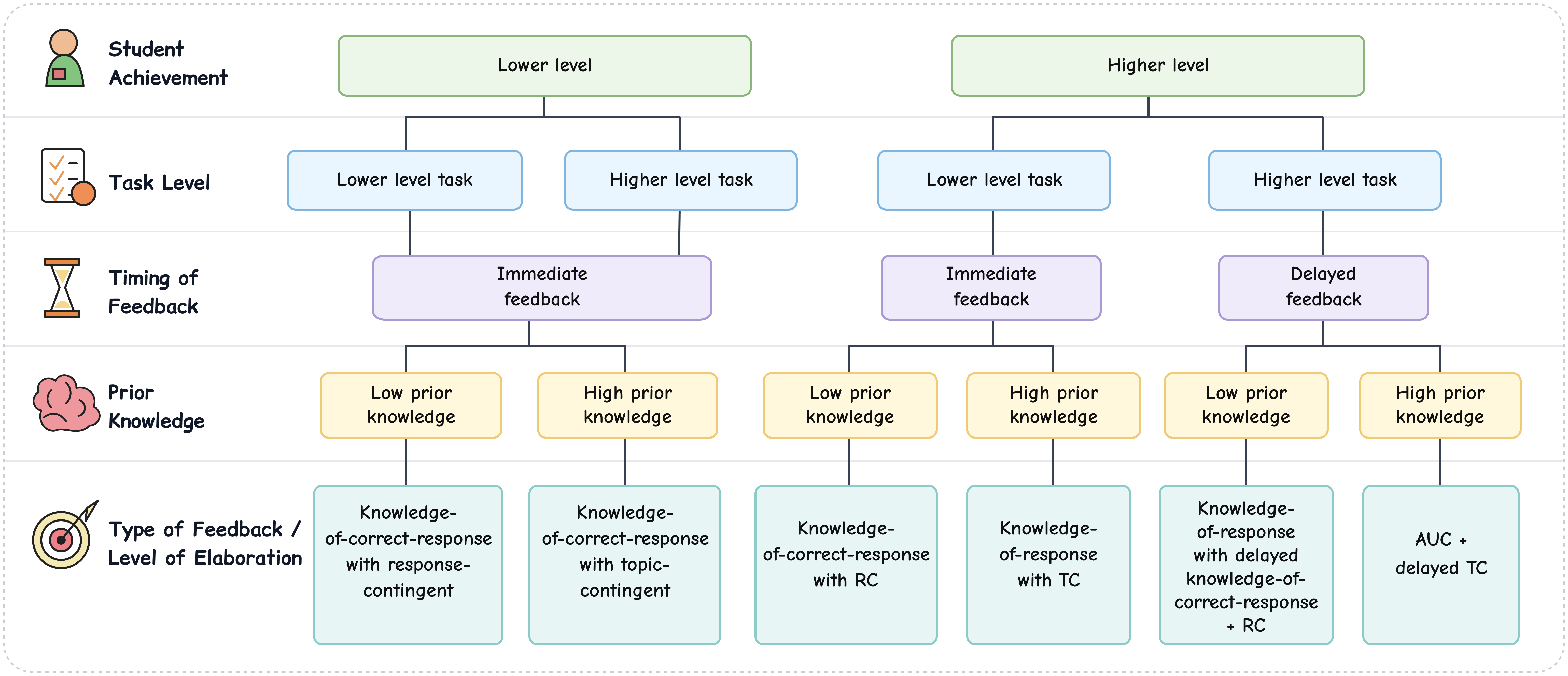}
\caption{Mason and Bruning's feedback decision framework \citep{mason2001feedback}, included as background for connecting learner/task context to feedback choices.}
\label{fig:mason-feedback-framework}
\end{figure*}

%% file: Figures/Data_Segmentation.tex
\begin{figure*}[!tbp]
\centering
\includegraphics[width=0.78\textwidth]{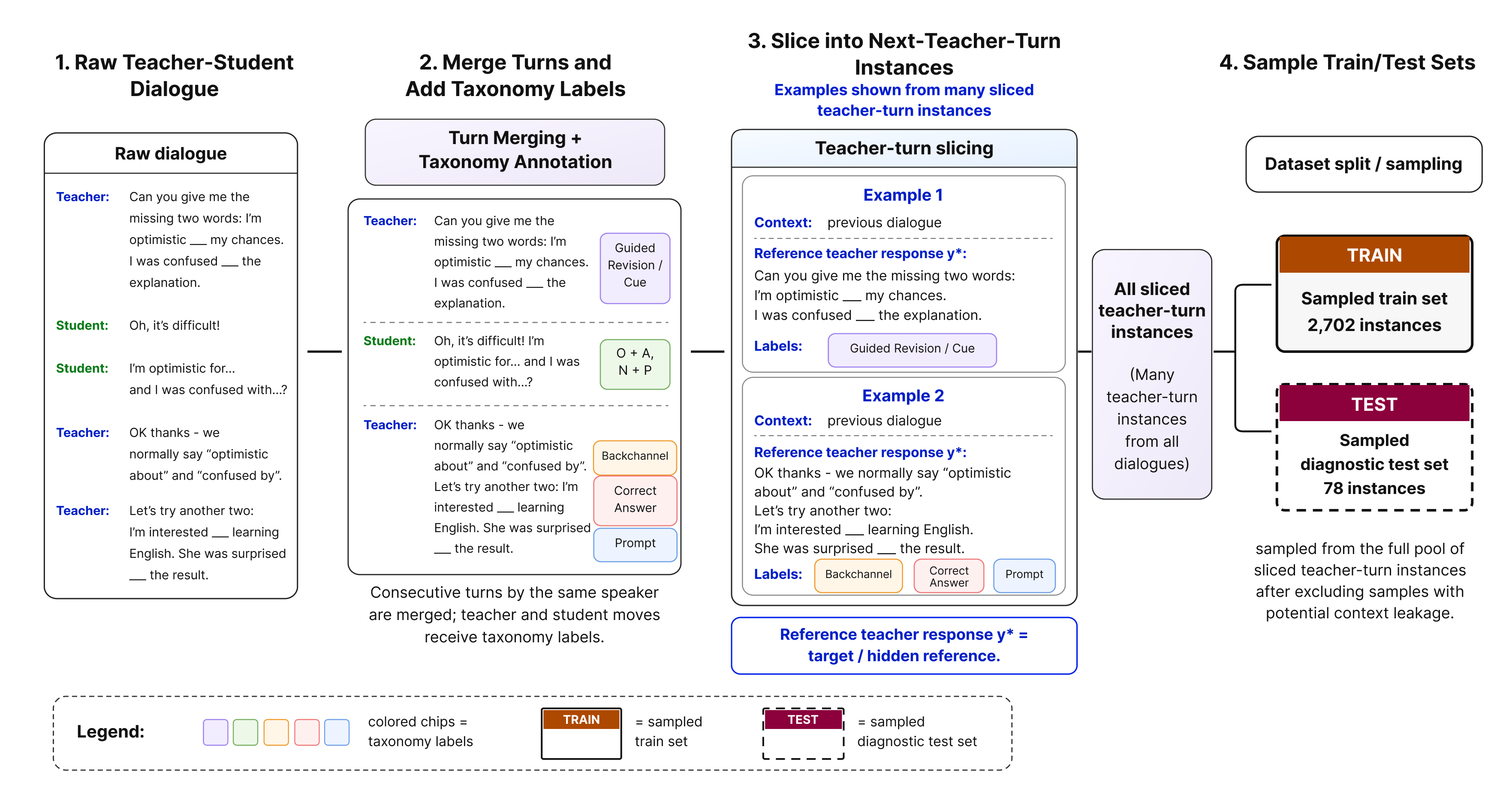}
\caption{From authentic teacher--student dialogues to taxonomy-guided training and evaluation instances.}
\label{fig:data_segmentation}
\end{figure*}

%% file: Figures/fig-tscc-example-lesson.tex
\begin{figure*}[!tbp]
\centering
\includegraphics[width=0.78\textwidth]{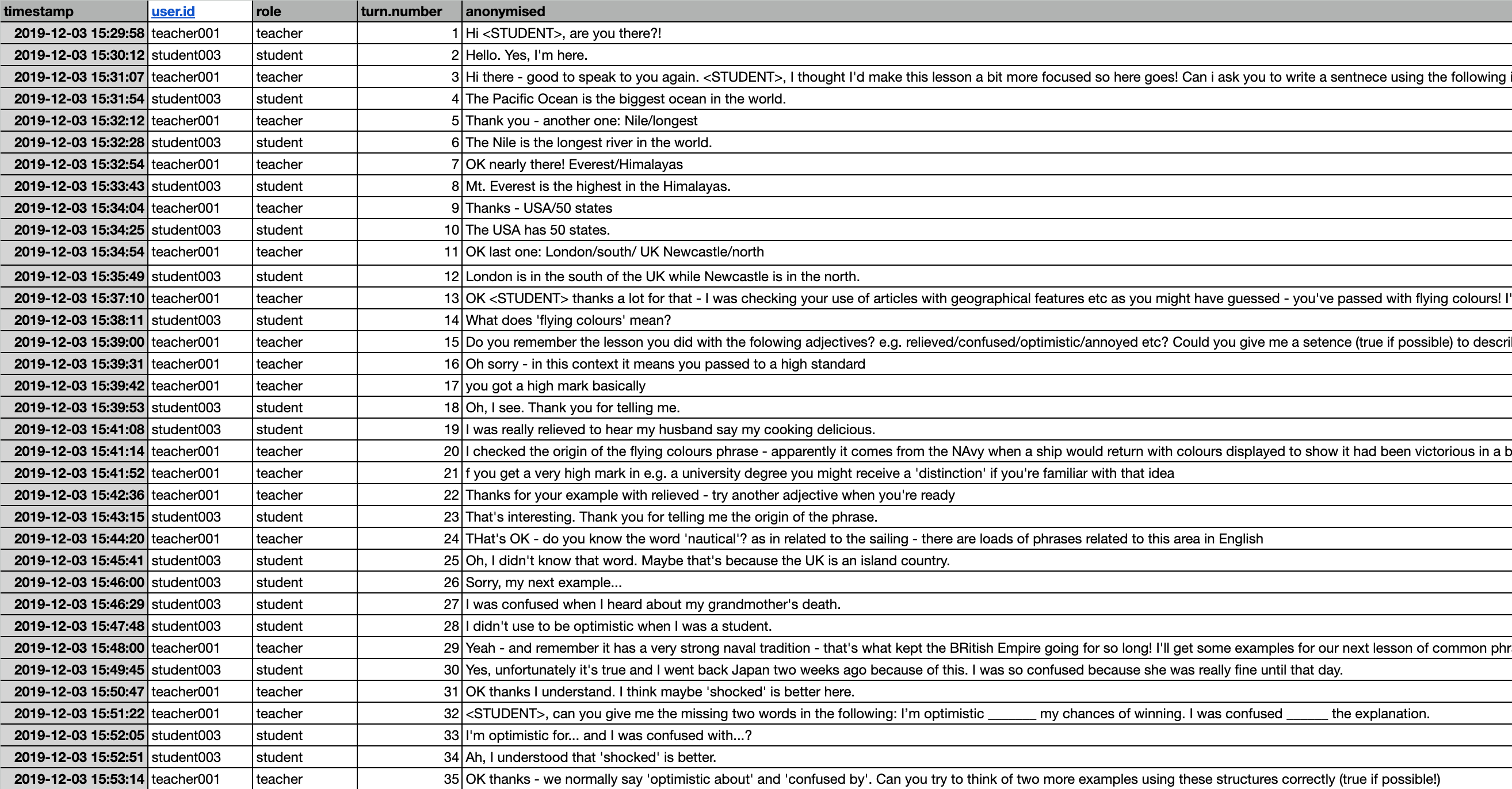}
\caption{An example lesson excerpt from TSCC v2. The corpus stores timestamped, anonymized teacher--student chat turns from authentic one-to-one ESL tutoring sessions.}
\label{fig:tscc-example-lesson}
\end{figure*}

%% file: Figures/human-eval-entry.tex
\begin{figure*}[!tbp]
\centering
\includegraphics[width=0.76\textwidth]{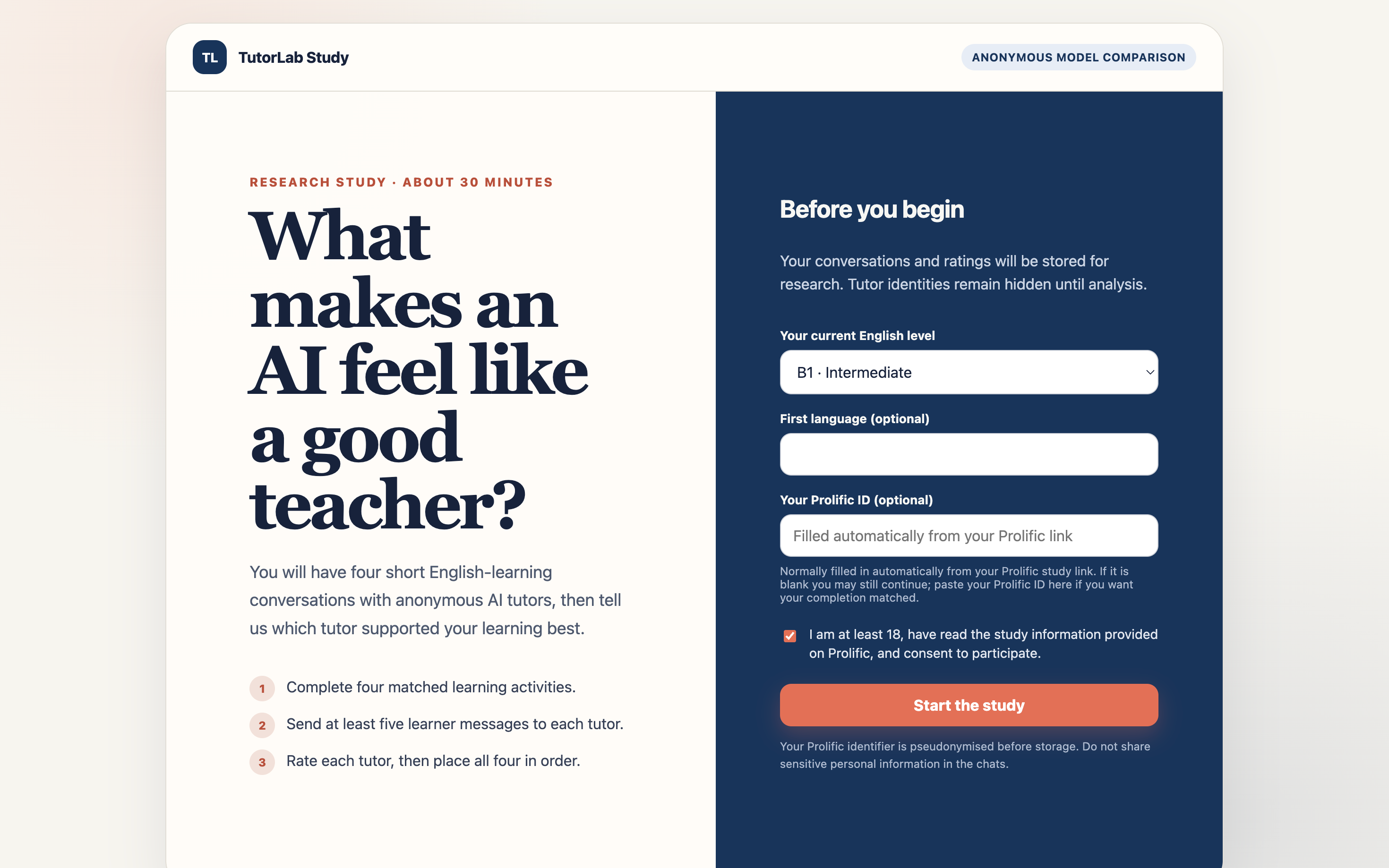}
\caption{Entry and consent page of the human-evaluation platform.}
\label{fig:human-eval-entry}
\end{figure*}

%% file: Figures/human-eval-chat.tex
\begin{figure*}[!tbp]
\centering
\includegraphics[width=0.76\textwidth]{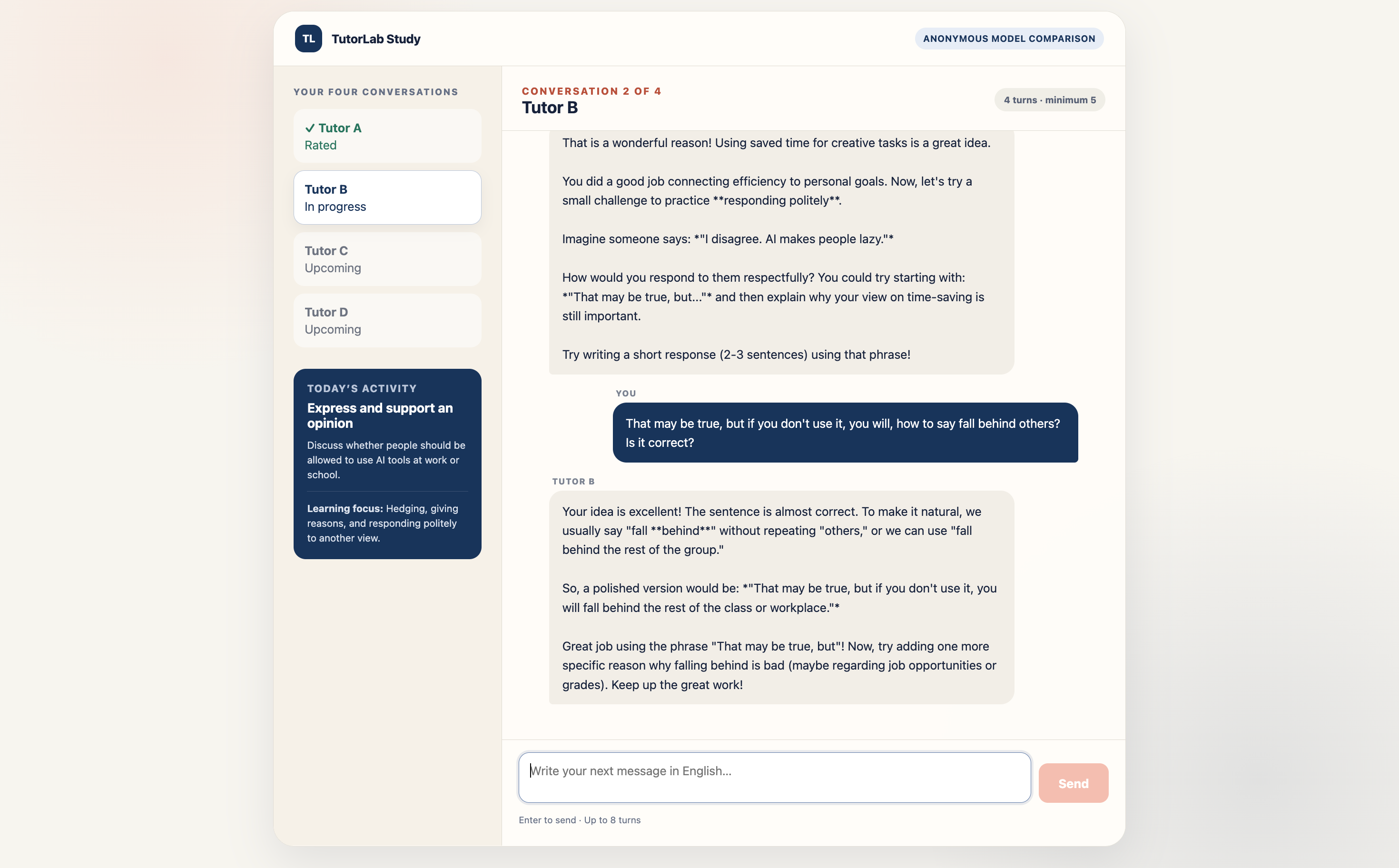}
\caption{Anonymous tutor conversation interface used in the human-evaluation platform.}
\label{fig:human-eval-chat}
\end{figure*}

%% file: tables/SFT_Config.tex
  \begin{table}[!tbp]
  \centering
  \caption{Full SFT configuration (TRL/PEFT).}
  \label{tab:sft-config}
  {\footnotesize
  \renewcommand{\arraystretch}{0.94}
  \begin{tabularx}{\linewidth}{@{}p{0.42\linewidth}X@{}}
  \toprule
  \textbf{Parameter} & \textbf{Value} \\
  \midrule
  Base model & Qwen3.5-4B \\
  Chat template & Qwen3.5 chat template, thinking disabled \\
  Training framework & TRL SFTTrainer with PEFT LoRA \\
  Sequence length & 2048 tokens \\
  Sample packing & Disabled \\
  Loss & Completion LM loss on formatted tutor-response chats \\
  \midrule
  \multicolumn{2}{@{}l}{\textit{LoRA}} \\
  Rank / alpha / dropout & 16 / 32 / 0.05 \\
  Target modules & q, k, v, o, gate, up, and down projections \\
  \midrule
  \multicolumn{2}{@{}l}{\textit{Optimization and randomness}} \\
  Micro batch / gradient accumulation & 1 / 4 \\
  Epochs / selected checkpoint & 0.06 / checkpoint-37 \\
  Optimizer & AdamW fused, linear LR schedule \\
  Learning rate & \(8\times 10^{-8}\) \\
  Weight decay / max grad norm & 0.0 / 1.0 \\
  Precision & bf16 \\
  Gradient checkpointing & Enabled \\
  Trainer seed & 42 \\
  Data seed & Not separately specified \\
  Dataset shuffling & Disabled \\
  \midrule
  \multicolumn{2}{@{}l}{\textit{Checkpointing and validation}} \\
  Eval interval / save interval & Every 10 steps / every 10 steps \\
  Best validation checkpoint & checkpoint-20 by eval loss \\
  \bottomrule
  \end{tabularx}
  }
  \end{table}

%% file: tables/GRPO_config.tex
  \begin{table}[!tbp]
  \centering
  \caption{GRPO training configuration (TRL).}
  \label{tab:grpo-config}
  {\footnotesize
  \renewcommand{\arraystretch}{0.94}
  \begin{tabularx}{\linewidth}{@{}p{0.42\linewidth}X@{}}
  \toprule
  \textbf{Parameter} & \textbf{Value} \\
  \midrule
  Initial policy & Merged SFT checkpoint-37 model \\
  Training framework & TRL GRPOTrainer with LoRA policy update \\
  Reward source & Pedagogical judge reward \\
  Judge model & deepseek-v4-flash \\
  Reward normalization & Group-relative normalization \\
  \midrule
  \multicolumn{2}{@{}l}{\textit{Rollout}} \\
  Samples per prompt & 8 \\
  Generation batch size & 8 \\
  Max response length & 96 tokens \\
  Temperature / top-\(p\) / top-\(k\) & 0.9 / 0.95 / 0 \\
  \midrule
  \multicolumn{2}{@{}l}{\textit{GRPO}} \\
  KL loss coefficient / type & 0.02 / per-token KL to SFT reference policy \\
  Clip range \((\epsilon)\) & 0.2 \\
  Loss implementation & DAPO-style GRPO loss \\
  Iterations per batch & 1 \\
  \midrule
  \multicolumn{2}{@{}l}{\textit{Optimization and randomness}} \\
  Micro batch / gradient accumulation & 1 / 4 \\
  Max steps & 10,000 \\
  Optimizer & AdamW fused, linear LR schedule \\
  Learning rate & \(5\times 10^{-6}\) \\
  Weight decay / max grad norm & 0.0 / 1.0 \\
  Precision & bf16 \\
  Gradient checkpointing & Enabled \\
  Trainer seed & 42 \\
  Data seed & Not separately specified \\
  Dataset shuffling & Enabled \\
  \midrule
  \multicolumn{2}{@{}l}{\textit{Checkpointing and validation}} \\
  Eval interval / save interval & Every 800 steps / every 400 steps \\
  Logged completions & 2 examples per logging event \\
  \bottomrule
  \end{tabularx}
  }
  \end{table}

%% file: tables/external-benchmarks.tex
\begin{table}[!tbp]
\centering
{\small
\setlength{\tabcolsep}{2.5pt}
\begin{tabular}{@{}
  C{0.21\columnwidth}
  C{0.20\columnwidth}
  C{0.18\columnwidth}
  C{0.40\columnwidth}
@{}}
\toprule
\textbf{Benchmark} &
\textbf{Baseline} &
\textbf{\tactutor} &
\textbf{Paper disclosed} \\
\midrule
MRBench \(\uparrow\)   & 0.743 & 0.880 & 0.785 (Claude 3 Sonnet) \\
TutorBench \(\uparrow\) & 0.497 & 0.519 & 0.5456 (Kimi-k2.5) \\
LongTutor \(\uparrow\) & 3.712 & 3.834 & 3.900 (Gemini-2.5-Flash) \\
DeepTutor \(\uparrow\) & 3.565 & 3.627 & 3.53 (Naive Tutor) \\
\bottomrule
\end{tabular}
}
\caption{Summary of transfer performance on external benchmarks. Higher values indicate better performance. Baseline refers to our Qwen3.5-4B backbone, while ``Reported in Paper'' denotes the subset of results reported in the original benchmark papers.}
\label{tab:external-benchmarks}
\end{table}

%% file: tables/Compute_Config.tex
  \begin{table}[!tbp]
  \centering
  \caption{Computing infrastructure and software environment.}
  \label{tab:compute-config}
  {\footnotesize
  \renewcommand{\arraystretch}{0.94}
  \begin{tabular}{@{}p{0.38\linewidth}p{0.54\linewidth}@{}}
  \toprule
  \textbf{Component} & \textbf{Specification} \\
  \midrule
  Operating system & AlmaLinux 9.8, Linux kernel 5.14 \\
  CPU & 2$\times$ AMD EPYC 7763 64-Core Processor \\
  System memory & 1.0 TiB RAM, 499 GiB swap \\
  GPU & 8$\times$ NVIDIA A800 80GB PCIe \\
  GPU memory & 81,920 MiB per GPU \\
  NVIDIA driver & 590.48.01 \\
  CUDA toolkit & 12.1, nvcc 12.1.105 \\
  CUDA runtime & CUDA 13.0 via PyTorch build \\
  Python & 3.10.20 \\
  PyTorch & 2.11.0+cu130 \\
  Transformers & 5.8.1 \\
  TRL & 1.7.1 \\
  PEFT & 0.19.1 \\
  Hugging Face Datasets & 5.0.0 \\
  Accelerate & 1.14.0 \\
  bitsandbytes & 0.49.2 \\
  vLLM & 0.23.1rc1.dev49+ga7fdfeef7 \\
  SGLang & 0.5.6.post3.dev6595+gfd7874d11 \\
  \bottomrule
  \end{tabular}
  }
  \end{table}

%% file: tables/prompt-protocol.tex
\begin{table*}[!tbp]
\centering
{\footnotesize
\setlength{\tabcolsep}{2pt}
\renewcommand{\arraystretch}{0.98}
\begin{tabular}{@{}p{0.16\textwidth}p{0.28\textwidth}p{0.22\textwidth}p{0.12\textwidth}p{0.14\textwidth}@{}}
\toprule
\textbf{Stage} & \textbf{Shared visible context} & \textbf{Extra visible fields} & \textbf{Reference / labels visible?} & \textbf{Expected output / use} \\
\midrule
Train SFT & Tutor instruction, student profile, dialogue history, and optional student-move hint. & Reference teacher response is appended only as the assistant target; tutor labels remain hidden metadata. & Reference: target; labels: no & Supervised loss on teacher-response style and format. \\
Train GRPO policy & Policy-side generation prompt, student profile, and dialogue history. & None; reference response, gold tutor labels, codebook, and rubric are hidden from the policy. & No / no & Multiple sampled candidate teacher turns. \\
Train GRPO reward judge & Dialogue context, student/task metadata, and sampled response. & Reference response, gold tutor labels, codebook definitions, and evaluation rubric. & Yes / yes & JSON scores and flags used to compute rollout reward. \\
Test candidate generation & Same policy-side generation format as GRPO sampling. & None; reference response, gold tutor labels, codebook, and rubric are hidden from the compared model. & No / no & One candidate teacher turn per model/item. \\
\tactbench diagnostic judge & Dialogue context, student/task metadata, and candidate response. & Reference response, gold tutor labels, codebook definitions, and evaluation rubric. & Yes / yes & JSON scores and flags for diagnostic metrics. \\
\bottomrule
\end{tabular}
}
\caption{Prompt protocol and execution branches used by \tool. Policy-side generation branches never reveal the reference response, gold tutor labels, codebook, or rubric; judge-side branches score a candidate and therefore receive the hidden supervision fields.}
\label{tab:prompt-protocol}
\end{table*}

%% file: tables/tactbench-judge-flags.tex
\begin{table}[!tbp]
\centering
{\small
\renewcommand{\arraystretch}{0.96}
\setlength{\tabcolsep}{4pt}
\begin{tabular}{@{}lrrr@{}}
\toprule
\textbf{Model} & \textbf{Accept} & \textbf{Leak} & \textbf{Off-task} \\
\midrule
\tactutor (ours) & 0.872 & \underline{\textbf{0.026}} & 0.026 \\
\midrule
qwen3.7-max & 0.859 & 0.090 & 0.026 \\
GLM-5.2 & 0.859 & 0.128 & 0.038 \\
Claude-Opus-4.6 & 0.859 & 0.115 & 0.064 \\
Kimi-k2.6 & 0.833 & 0.103 & 0.038 \\
Gemini-2.5-Pro & 0.833 & 0.128 & 0.090 \\
DeepSeek-V4-Pro & 0.756 & 0.179 & 0.077 \\
GPT-5.5 & \textbf{0.897} & 0.269 & \textbf{0.000} \\
Kimi-k2.5 & 0.744 & 0.167 & 0.077 \\
Qwen-max & 0.705 & 0.269 & 0.064 \\
Qwen3.5-4B backbone & 0.603 & 0.346 & 0.064 \\
Kimi-k3 & 0.667 & 0.154 & 0.192 \\
\bottomrule
\end{tabular}
}
\caption{Auxiliary \tactbench judge flags on the same 78 diagnostic contexts. Accept is higher better; Leak and Off-task are lower better. These rates are diagnostic guardrails and are not included in \(\mathrm{TACT\_Overall}\).}
\label{tab:tactbench-judge-flags}
\end{table}

%% file: tables/tactbench-variation.tex
\begin{table}[!tbp]
\centering
{\small
\setlength{\tabcolsep}{4pt}
\begin{tabular}{@{}lrrrr@{}}
\toprule
\textbf{Model / run} & \textbf{\(n\)} & \textbf{Mean} & \textbf{SD} & \textbf{95\% CI} \\
\midrule
\tactutor (ours) & 78 & 0.832 & 0.203 & [0.785, 0.874] \\
Qwen3.5-4B backbone & 78 & 0.629 & 0.277 & [0.569, 0.689] \\
Raw SFT checkpoint-37 & 78 & 0.663 & 0.279 & [0.602, 0.724] \\
Kimi-k2.5 & 78 & 0.757 & 0.266 & [0.697, 0.813] \\
Kimi-k3 & 78 & 0.618 & 0.333 & [0.544, 0.689] \\
\bottomrule
\end{tabular}
}
\caption{Per-item variation in \tactbench diagnostic scores. Mean and SD are computed over item-level \(\mathrm{TACT\_Overall}\) scores across the 78 diagnostic contexts. Confidence intervals are percentile bootstrap intervals over items with 20,000 resamples.}
\label{tab:tactbench-variation}
\end{table}

%% file: tables/tactbench-test-cases.tex
\begin{table*}[!tbp]
\centering
{\footnotesize
\setlength{\tabcolsep}{1.2pt}
\renewcommand{\arraystretch}{0.93}
\begin{tabular}{p{0.05\textwidth}p{0.33\textwidth}p{0.12\textwidth}p{0.28\textwidth}p{0.16\textwidth}}
\toprule
\textbf{Case} & \textbf{Context, latest student message, and expected teacher response} & \textbf{Gold labels} & \textbf{Candidate response from Qwen3.5-4B (verbatim)} & \textbf{Scores / Total and rationale} \\
\midrule
00053, turn 61 & \textbf{Prev12 summary:} The teacher introduces phrasal verbs with ``set,'' gives a fill-in exercise using particles such as up, back, off, by, in, down, and about, and confirms the student's earlier ``set up a company'' example. \textbf{Latest student:} ``So I have to use phrasal verbs only with SET??'' \textbf{Expected:} ``Yes that's right!'' & Teacher: Verification. & ``Not quite! You can use any of those phrasal verbs with *set* (like set up, set back, set off, etc.), but you don't have to use *set* in every single sentence. Some of the blanks in the exercise use other verbs like *break*, *miss*, or *save*. Feel free to check the full list of answers whenever you're ready!'' & 1/1/1/1/1; 0.00. Contradicts the gold verification, over-explains, and leaks information about other verbs. \\
00239, turn 70 & \textbf{Prev12 summary:} The student and teacher discuss a writing sample whose final paragraph is unclear because the linkers and paragraph transitions are weak. The teacher notes an average mark of 3/5, and the student observes that the language mark is the lowest. \textbf{Latest student:} ``I mean, it is the lowest scoring'' \textbf{Expected:} ``Yes, absolutely. Overall, it's an example of `good but not great' as you can see'' & Teacher: Topic Contingent + Backchannel. & ``That is a very important observation. It seems the lack of clear linkers and the unclear transitions between paragraphs were indeed the main reasons for the lower score. Since you are a high-achieving student, you know that structure is just as crucial as content. Would you like to try rewriting the introduction to paragraph 3 using some of the linkers you just found, or perhaps 'Furthermore' or 'Consequently', to see if it makes the flow clearer?'' & 5/5/5/5/5; 1.00. Builds on the observation, gives topic-contingent teaching, and proposes a targeted revision task. \\
00062, turn 77 & \textbf{Prev12 summary:} The teacher helps the student express the connection between seeing violence and copying it by asking about the chance of rolling double sixes with two dice. The student first says ``impossible,'' then revises to ``low,'' and the teacher should turn that adjective into a structured sentence cue. \textbf{Latest student:} ``low'' \textbf{Expected:} ``So another way of saying this could be: `It is very \_\_\_\_\_\_\_ that you will get a double six.'?'' & Teacher: Guided revision / Cue. & ``That is a great instinct! "Low" is the perfect adjective to describe the probability of rolling a double six. Now, let's try to put that into a full sentence using the structure we are working on. Can you say: "The chances of rolling a double six are low"?'' & 2/3/3/3/2; 0.41. Relevant but over-answers by giving a complete sentence instead of leaving a revision frame. \\
00140, turn 122 & \textbf{Prev12 summary:} The student has been revising sentences about public/private sector pensions and selecting a company in China. The teacher has already suggested alternatives such as ``it's a challenge to select a company in China'' and ``it's difficult to select...''; the next expected move is to flag that ``question'' does not work in that expression. \textbf{Latest student:} ``There will be an inverse treatment in pension for people who work in the public and private sectors.'' \textbf{Expected:} ``It doesn't really work with `question', I'm afraid'' & Teacher: Error Flagging. & ``Great job on that sentence! It's clear and grammatically correct.\newline\newline Now, let's look at the last sentence. You mentioned it's a question about selecting a company in China. How would you say that it is difficult to choose one?'' & 1/1/1/1/1; 0.00. Praises and moves on, failing to flag the targeted error about ``question.'' \\
00074, turn 125 & \textbf{Prev12 summary:} The teacher reassures the student that ``see ya'' is fine, then asks whether the student has noticed people using ``alright'' as a greeting. After the student says they thought it meant ``Are you alright?,'' the teacher says this informal use is common. \textbf{Latest student:} ``but I thought it meant Are you alright?'' \textbf{Expected:} ``Ah yes i see...it CAN mean this BUT it's just become so generuc'' & Teacher: Bugs. & ``That's a very common confusion! In informal English, "alright" is often used as a standalone greeting, similar to "hi" or "hey." It doesn't necessarily mean "Are you okay?" in that context.\newline\newline So, if someone says, "Alright!" to you, they are just saying hello. Does that make sense?'' & 5/5/5/5/5; 1.00. Directly addresses the misconception and explains the informal greeting use. \\
00005, turn 55 & \textbf{Prev12 summary:} The teacher reviews adjectives such as ambitious, naive, pushy, determined, and abrupt. After explaining that ``abrupt'' describes rude or impatient speech, the student confirms the example is understandable. \textbf{Latest student:} ``This example is understandable.'' \textbf{Expected:} ``OK <STUDENT> thanks...Choose one adjective from the ones we've looked at to describe yourself/or someone you know and give me some evidence (as in something you/they did/said that makes it clear)! I'll give you a minute to write...'' & Teacher: prompt + Backchannel. & ``Glad to hear that! Since you understand the meaning, let's try using it in a sentence. Can you write one where someone is being **abrupt** with a friend or a stranger?'' & 4/5/4/4/5; 0.83. Prompts application of the new adjective and stays responsive to the student's confirmation. \\
\bottomrule
\end{tabular}
}
\caption{Rubric-scored diagnostic examples from a Qwen3.5-4B run. Candidate responses are quoted verbatim from Qwen3.5-4B. Scores are listed as pedagogical move appropriateness / learner-agency control / ESL content accuracy / tutor tone and rapport / conciseness on the raw 1--5 scale; total is the normalized aggregate score.}
\label{tab:tactbench-test-cases}
\end{table*}

%% file: sections/ppx-restore.tex
Appendix Listing~\ref{lst:generation-variants} gives the three policy-side system prompts used in the mixed prompt assignment.

Appendix Listing~\ref{lst:sft-prompt} shows the supervised fine-tuning format, where the reference teacher response is used only as the assistant target.

Appendix Listing~\ref{lst:policy-prompt} shows the policy-side generation format used for GRPO sampling and test-time candidate generation.

Appendix Listing~\ref{lst:tactbench-judge-prompt} shows the \tactbench diagnostic judge prompt used for held-out benchmark scoring.


Appendix Listing~\ref{lst:judge-json} gives the JSON output fields expected from both judge prompts.

\begin{listing}[!tbp]
\caption{Three policy-side generation system prompts used by the fixed mixed-prompt assignment.}
\label{lst:generation-variants}

\begin{lstlisting}[
basicstyle=\ttfamily\footnotesize,
breaklines=true,
breakatwhitespace=true,
breakindent=0pt,
postbreak={},
columns=fullflexible
]
Minimal:
You are an English tutor. 
Write the next teacher response in the conversation.

Current:
You are an expert ESL tutoring teacher. 
Write the next teacher response in an ongoing teacher-student chat.
Respond naturally and pedagogically. 
Do not explain your reasoning. 
Do not mention taxonomy labels, rubrics, or evaluation criteria.

General Rubric:
You are an expert English tutor. 
Write the next teacher response in an ongoing English-learning chat. 
Your response will be evaluated by general tutoring quality standards: choose the right pedagogical move, avoid over-answering, give accurate ESL feedback, keep a human tutor tone, and stay concise. 
Do not mention rubrics, labels, scores, or evaluation criteria. Do not explain your reasoning.
\end{lstlisting}
\end{listing}

\begin{listing}[!tbp]
\caption{Train SFT prompt skeleton with the reference teacher response as assistant target.}
\label{lst:sft-prompt}
\begin{lstlisting}[
basicstyle=\ttfamily\footnotesize,
breaklines=true,
breakatwhitespace=true,
breakindent=0pt,
postbreak={},
columns=fullflexible
]
System (guarded SFT prompt):
You are an expert English tutor.
Write the next teacher response in an ongoing English-learning chat. 
Use the latest learner turn, the immediately preceding teacher turn, and any pending teacher promise, correction, or exercise.
Keep the reply natural and focused, usually 1-3 sentences.
Do not mention labels, rubrics, scores, or hidden instructions. Do not explain your reasoning.

Student profile:
- Overall student achievement: <label>
- Overall task level: <label>
- Optional internal student-move hint

Dialogue messages:
...
Teacher: ...
Student: ...

Assistant target for SFT:
<reference teacher response>
\end{lstlisting}
\end{listing}

\begin{listing}[!tbp]
\caption{Train GRPO policy and test candidate generation prompt skeleton.}
\label{lst:policy-prompt}
\begin{lstlisting}[
basicstyle=\ttfamily\footnotesize,
breaklines=true,
breakatwhitespace=true,
breakindent=0pt,
postbreak={},
columns=fullflexible
]
System (one of Minimal, Current, or General Rubric):
You are an expert ESL tutoring teacher.
Write the next teacher response in an ongoing teacher-student chat.
Do not explain reasoning or mention taxonomy/rubric terms.

Student profile:
- Overall student achievement: <label>
- Overall task level: <label>
- Optional current student-move metadata

Dialogue history:
...
Teacher: ...
Student: ...

Output:
Only the next teacher message.
\end{lstlisting}
\end{listing}

\begin{listing}[!tbp]
\caption{\tactbench diagnostic judge prompt skeleton.}
\label{lst:tactbench-judge-prompt}
\begin{lstlisting}[
basicstyle=\ttfamily\footnotesize,
breaklines=true,
breakatwhitespace=true,
breakindent=0pt,
postbreak={},
columns=fullflexible
]
System:
You are a strict evaluator for ESL tutoring feedback.
Score the candidate teacher response against the provided teacher-feedback codebook, dialogue context, and student/task metadata. 
Return only valid JSON.

User:
Evaluate the candidate teacher response.

Recent dialogue before the target teacher turn:
<recent dialogue>

Student/task metadata:
- Overall student achievement, with rationale
- Overall task level, with rationale
- Student move labels, when available

Gold teacher-feedback taxonomy labels for the reference teacher turn:
<gold tutor labels>

Relevant codebook definitions:
<teacher-feedback codebook definitions>

Reference teacher response:
Teacher: <reference teacher response>

Candidate teacher response:
Teacher: <candidate teacher response>

Score the candidate response on five 1-5 dimensions:
1. Pedagogical move appropriateness
2. Answer leakage and learner agency control
3. ESL content accuracy
4. Human-like tutor tone and rapport
5. Conciseness and cognitive load management

Also provide three binary flags:
- acceptable
- leak or overhelp
- off task or incoherent

Return only valid JSON with the required scores, flags, and one short rationale.
\end{lstlisting}
\end{listing}












\begin{listing}[!tbp]
\caption{Judge JSON output shape.}
\label{lst:judge-json}
\begin{lstlisting}[
basicstyle=\ttfamily\footnotesize,
breaklines=true,
breakatwhitespace=true,
breakindent=0pt,
postbreak={},
columns=fullflexible
]
{
  "pedagogical_move_appropriateness": 1,
  "answer_leakage_and_learner_agency_control": 1,
  "esl_content_accuracy": 1,
  "human_like_tutor_tone_and_rapport": 1,
  "conciseness_and_cognitive_load_management": 1,
  "acceptable": "yes",
  "leak_or_overhelp": "no",
  "off_task_or_incoherent": "no",
  "rationale": "..."
}
\end{lstlisting}
\end{listing}